\documentclass[nohyperref]{article}
\pdfoutput=1
\usepackage{microtype}
\usepackage{graphicx}
\usepackage{booktabs} % for professional tables

\usepackage{hyperref}
\usepackage{caption}
\usepackage{subcaption}

\usepackage[accepted]{icml2022}

\usepackage{amsmath}
\usepackage{amssymb}
\usepackage{mathtools}
\usepackage{amsthm}
\usepackage{color}

\usepackage[capitalize,noabbrev]{cleveref}

\DeclareMathOperator*{\argmax}{arg\,max}

\theoremstyle{plain}

\theoremstyle{definition}

\theoremstyle{remark}

\usepackage[textsize=tiny]{todonotes}
\usepackage{verbatim}

\icmltitlerunning{Submission and Formatting Instructions for ICML 2022}

\def\tcr{\textcolor{red}}

\begin{document}

\twocolumn[
\icmltitle{MASER: Multi-Agent Reinforcement Learning with Subgoals Generated from Experience Replay Buffer}

% It is OKAY to include author information, even for blind
% submissions: the style file will automatically remove it for you
% unless you've provided the [accepted] option to the icml2022
% package.

% List of affiliations: The first argument should be a (short)
% identifier you will use later to specify author affiliations
% Academic affiliations should list Department, University, City, Region, Country
% Industry affiliations should list Company, City, Region, Country

% You can specify symbols, otherwise they are numbered in order.
% Ideally, you should not use this facility. Affiliations will be numbered
% in order of appearance and this is the preferred way.
\icmlsetsymbol{equal}{*}
\icmlsetsymbol{equalAdv}{$\dagger$}

\begin{icmlauthorlist}
\icmlauthor{Jeewon Jeon}{yyy}
\icmlauthor{Woojun Kim}{yyy}
\icmlauthor{Whiyoung Jung}{yyy}
\icmlauthor{Youngchul Sung}{yyy}
%\icmlauthor{Firstname5 Lastname5}{yyy}
%\icmlauthor{Firstname6 Lastname6}{sch,yyy,comp}
%\icmlauthor{Firstname7 Lastname7}{comp}
%\icmlauthor{}{sch}
%\icmlauthor{Firstname8 Lastname8}{sch}
%\icmlauthor{Firstname8 Lastname8}{yyy,comp}
%\icmlauthor{}{sch}
%\icmlauthor{}{sch}
\end{icmlauthorlist}

\icmlaffiliation{yyy}{School of Electrical Engineering, KAIST, Daejeon, South Korea}
%\icmlaffiliation{comp}{Company Name, Location, Country}
%\icmlaffiliation{sch}{School of ZZZ, Institute of WWW, Location, Country}

\icmlcorrespondingauthor{Youngchul Sung}{ycsung@kaist.ac.kr}
\icmlcorrespondingauthor{Woojun Kim}{woojun.kim@kaist.ac.kr}

% You may provide any keywords that you
% find helpful for describing your paper; these are used to populate
% the "keywords" metadata in the PDF but will not be shown in the document
\icmlkeywords{Machine Learning, ICML}

\vskip 0.3in
]

% this must go after the closing bracket ] following \twocolumn[ ...

% This command actually creates the footnote in the first column
% listing the affiliations and the copyright notice.
% The command takes one argument, which is text to display at the start of the footnote.
% The \icmlEqualContribution command is standard text for equal contribution.
% Remove it (just {}) if you do not need this facility.

\printAffiliationsAndNotice{}  % leave blank if no need to mention equal contribution
%\printAffiliationsAndNotice{\icmlEqualContribution} % otherwise use the standard text.

\begin{abstract}
In this paper, we consider cooperative multi-agent reinforcement learning (MARL) with sparse reward. To tackle this problem, we propose a novel method named MASER: MARL with  subgoals generated from experience replay buffer.  Under the widely-used assumption of centralized training with decentralized execution and consistent Q-value  decomposition for MARL,
MASER automatically generates proper subgoals for multiple agents from the experience replay buffer by considering both individual Q-value and  total Q-value. Then, MASER designs individual  intrinsic reward for each agent based on actionable representation  relevant to Q-learning so that the agents reach their subgoals while maximizing the joint action value. Numerical results show that MASER significantly outperforms StarCraft II  micromanagement benchmark compared to other state-of-the-art MARL algorithms.
\end{abstract}

\section{Introduction}
%%%%%%%%%%%%%%%%%%%%%%%%%%%%%%%%%%%%%%%%%%

\begin{comment}
\begin{itemize}
    \item MARL
    \item Sparse
    \item Subgoal and intrinsic reward:         Subgoal: modeling and roll out and limitation. So, we use experience replay buffer
        \begin{itemize}
        \item existing method's problem
    \end{itemize}
        \item Our method: 
\end{itemize}
\end{comment}

\begin{comment}
Deep Reinforcement Learning (RL) has been demonstrated to effectively learn a wide range of real-world tasks including online games \cite{online_game1, online_game2}, robotics \cite{robotic1, robotic2}, and autonomous vehicles \cite{autonomous_car}. 
\end{comment}

Deep multi-agent reinforcement learning (MARL) is an active research field  to handle many real-world problems such as  games, traffic control, and connected self-driving cars, which can be  modeled as multi-agent systems \cite{2019Li, 2019Andriotis}.
Since the introduction of the simplest approach of independent Q-learning (IQL) \cite{1993Tan}, many advanced MARL algorithms  have  been developed recently, e.g.,  VDN \cite{VDN}, QMIX \cite{QMIX}, QTRAN \cite{QTRAN}, COMA \cite{2018Foerster}, MADDPG \cite{MADDPG}, Social Influence \cite{2019Jaques}, Intention Sharing  \cite{MARL-IS}.
Although these recent algorithms effectively handle the increased dimensions  of joint state and action spaces associated with multiple agents  and alleviate the burden of inter-agent communication \cite{2019Oroojlooyjadid, 2018rashid}, they were primarily developed under the assumption of dense reward.   However, many multi-agent problems are modeled as sparse-reward environments in which a non-zero reward is not given to agents' actions in every time step but given only when certain conditions are met. For example, in multi-agent robot soccer an actual non-zero reward occurs only when scoring a goal, and individual actions such as passes, dribbles or tackles do not receive non-zero rewards.  Thus, non-zero rewards come very seldom in such sparse-reward environments. Reinforcement learning in sparse-reward environments is difficult in general and is more challenging in multi-agent environments since a global reward is shared among multiple agents and the combination of multiple agents' actions jointly affects the state evolution in multi-agent systems.

A common approach to  tackle sparse-reward MARL is learning with efficient exploration \cite{cmae, seac, remax, maven, MMI}, which has been shown effective in some tasks. 
However, with exploration only, it is difficult to learn which action induces  rare non-zero rewards.  
As an alternative to direct learning of assigning  credit to trajectory, which is an essential difficulty in sparse environments,  
subgoal-based methods are recently emerging as an effective approach  to sparse-reward MARL  \cite{hdmarl, roma, rode},  inspired by the methods using subgoals in  single-agent systems \cite{hdqn, HER, generate_subgoal, imagined_subgoal}.  In subgoal-based methods, typically subgoals for maximizing the total episodic reward are determined or designed   and learning is performed to reach the subgoals to attain the final goal.   The key elements here are how to determine or design subgoals and how to learn to reach the subgoals.    
One way to determine subgoals is to select from a set of pre-defined subgoals designed based on domain knowledge for a given task \cite{hdmarl,hdqn}. However, such an approach is task-specific and not general enough to be applicable to a variety of different tasks. 
Thus, \emph{automatic subgoal generation} is desirable for subgoal-based approaches to be applicable to various tasks without being dependent on task-specific domain knowledge. 
Automatic subgoal generation has been considered in single-agent systems  \cite{HER, generate_subgoal, imagined_subgoal}. In multi-agent systems, however, automatic subgoal generation has not been investigated much yet. In addition to  subgoal generation, subgoal-based methods require a way to reach the subgoals, and reaching subgoals is  typically attained by employing intrinsic rewards. 
In single-agent systems, reaching the subgoals through intrinsic rewards is relatively easy  because a single intrinsic reward is designed in single-agent systems \cite{HER}.
In multi-agent systems, on the other hand, designing intrinsic rewards is challenging because we need to design multiple intrinsic rewards for multiple agents considering different contributions of different agents to achieve the final goal.  %\tcr{ remove \cite{LIIR, GIIR} has been proposed.} 

In order to deal with these difficulties in subgoal-based MARL, we propose a novel method named MASER: Multi-Agent reinforcement learning  with  Subgoals generated from Experience Replay buffer.  Under the widely-used assumption of centralized training with decentralized execution (CTDE) \cite{2019Oroojlooyjadid, 2018rashid} and consistent Q-value  decomposition for MARL \cite{QMIX},
MASER automatically generates proper subgoals for multiple agents from the experience replay buffer by compromising  the individual Q-values and the total Q-value. MASER designs local and global intrinsic rewards based on actionable representation  \cite{ARC} relevant to Q-learning so that the agents can reach their subgoals while maximizing the joint action value. Numerical results show that MASER significantly outperforms existing algorithms in MARL with sparse reward.

%%%%%%%%%%%%%%%%%%%%%%%%%%%%%
\section{Related Works}
%%%%%%%%%%%%%%%%%%%%%%%%%%%%%

% \subsection{Demonstration-guided RL}
% In the real-world, designing reward is challenging due to an enormous combination of reasons that affect rewards. To design reward efficiently, demonstration-guided RL which is an approach for learning behaviors from reward feedback \cite{demonstration_RL, DQFD, DDPGFD} has been utilized. They learn from the demonstration sample obtained by experts compared with the sample which they collect by exploration. To obtain the demonstration sample, the prior domain knowledge is demanded which is task-specific. 

\textbf{MARL with Value Decomposition:}  Value decomposition is one of the key techniques for MARL with finite action space, enabling CTDE, e.g., VDN \cite{VDN}, QMIX \cite{QMIX}, QTRAN \cite{QTRAN}. In QMIX,  the joint action-value function $Q^{tot}$ is decomposed into local Q-value functions associated with individual agents and then the overall Q-value  is obtained through a mixing network, where some constraint is applied to guarantee consistency between the local Q-values and the overall Q-value.  In this paper, we use a mixing network similar to that  introduced in QMIX.

\textbf{Subgoal Assignment:}
There exist previous works on subgoal-based MARL with sparse reward.  HDMARL \cite{hdmarl} proposed a hierarchical deep MARL scheme to solve sparse-reward tasks with temporal abstraction.
HDMARL  selects one  from a set of predefined subgoals designed based on  domain knowledge and hence such subgoal design is not easily applicable to different tasks. 
HSD \cite{hsd} trains cooperative decentralized policies at a high level to select skills, and the agents take actions conditioned by chosen skills from high-level.

In the single-agent case,  \citet{generate_subgoal} generated subgoal by restricting  the goal space based on adjacency constraint and \citet{imagined_subgoal} defined a subgoal set composed of the midpoints between the current state and the goal state and selected a subgoal from the subgoal set. \citet{HER} proposed using replay buffer to determine subgoals. However, this method simply chooses a state randomly  in a failed trajectory or the final state of a failed trajectory as the subgoal. On the other hand, MASER selects multiple effective subgoals for multiple agents from  experience episodes by considering their local and global Q-values so that the selected subgoals provide more credit to the final goal achievement.  

ROMA \cite{roma} and RODE \cite{rode} give roles to agents to decompose the overall task, where role is a bit different concept from subgoal. 
In particular, in ROMA, the role is learned by the agents and the agents with  similar roles perform similar sub-tasks to increase the overall task efficiency. In contrast to the roles in ROMA, the subgoals in our approach are determined from the entire trajectory and guide the multiple agents to their own goals. In addition, MASER takes advantage of using centralized training for determining subgoals, but ROMA determines the roles based on local observations of the agents.

\textbf{MARL with Individual Intrinsic Rewards:}
Individual intrinsic reward generation for multiple agents in MARL has been considered.   
By assigning individual reward for each agent, agents are stimulated differently.
LIIR \cite{LIIR} directly learns individual intrinsic reward for each agent to update a proxy critic to obtain each agent's policy, while intrinsic reward learning is updated to maximize extrinsic reward sum. 
Similarly to LIIR, GIIR \cite{GIIR} also learns individual intrinsic reward for each agent for multi-agent Q-learning. 
Unlike this direct intrinsic reward learning, our method generates individual intrinsic reward based on generated subgoals to attain the final goal.

\textbf{Representation Learning in RL:}
Representation learning in RL deals with the transformation of state or observation (i.e., partial state) into a form that captures information relevant to control \cite{mapping_state, laplacian}. Actionable representation for control (ARC) provides an effective way to capture the elements in the observation that are necessary for decision making \cite{ARC}. For this, \citet{ARC} defined the actionable distance between two goal states $s_1$ and $s_2$ via an intermediate state $s$ as the symmetrized KL divergence between two policies conditioned on two states: $\pi(a|s,s_1)$ and $\pi(a|s,s_2)$. This basically quantifies the difference in actions to reach two states $s_1$ and $s_2$ from $s$. When the required actions are totally different, the actionable distance is large. When the required actions are similar, on the other hand, the actionable distance is small.
Then, ARC is learned based on the actionable distance. We also use actionable representation to obtain the intrinsic reward from  partial states chosen as subgoals. In our case, to be compatible with multi-agent  Q-learning, we define our actionable distance using the local Q-function. This new actionable distance has a very simple form and is well suited to Q-learning.
%\tcr{This choice facilitates multi-agent learning XXXXX Proof related.}

\begin{comment}
\textbf{MARL with Exploration:}   Efficient exploration is used to solve MARL with sparse reward. CMAE \cite{cmae} proposed cooperative multi-agent exploration with sharing a common goal among agents. The goal is selected from projected state spaces instead of the whole state space via an entropy-based technique. SEAC \cite{seac} shares experience among agents to do diverse exploration. It considers other agents' trajectories as off-policy data to apply experience sharing by combining the gradients of different agents. MAVEN \cite{maven} introduces a shared latent variable which is controlled by a hierarchical policy.  The shared latent variable allows agents to achieve committed exploration.
\end{comment}

%%%%%%%%%%%%%%%%%%%%%%%%%%%%%%%%%%%%%%%%%%%%%%
\section{Background}
%%%%%%%%%%%%%%%%%%%%%%%%%%%%%%%%%%%%%%%%%%%%%%

\textbf{Decentralized Partially Observable Markov Decision Process:}  A cooperative multi-agent task with $N$ agents can be described as a decentralized partially observable Markov decision process (Dec-POMDP) \cite{Dec-POMDP}. A Dec-POMDP consists of a tuple $G = \langle S, U, P, Z, O, r, \gamma, N \rangle$,  where $S$ is the state space, $U$
is the action space, $P$ is the transition probability, $Z$ is the observation space, $O$ is the observation function, $r$ is the reward function, $\gamma$ is the discount factor and $N$ is the number of agents. The action space is assumed to be finite in this paper.  At each time step, each agent $i \in I := \{1, \dots, N \}$  chooses an action $u^i \in U$ and the joint action $\mathbf{u}:=(u^1,\cdots,u^N) \in U^N =: \mathbf{U}$ is formed based on the actions of all agents. Then, the joint action causes a transition in the environment state according to the state transition probability $P(s'|s, \mathbf{u}) : S \times \mathbf{U} \times S \rightarrow [0, 1]$, where $s$ is the global state of the environment and $s'$ is the next global state.  Due to partial observability,  Agent $i$ makes an individual observation $o^i \in Z$ following the observation function $O(s, i) : S \times I \rightarrow Z$ at each time step. 
The action $u^i$ of Agent $i$ is selected according to its policy $\pi^i(u^i|\tau^i)$, where  $\tau^i$ 
is the observation-action history of Agent $i$ up to the current time step.  The policies of all agents form a joint policy $\pi=(\pi^1,\cdots,\pi^N)$, and the goal is to maximize the expected global return $\mathbb{E}[G_0]$ by optimizing  $\pi$, where $G_t = \sum_{k=0}^\infty \gamma^k r_{t+k}$ is the discounted return and $r_t$ is the global reward at time step $t$  according to the reward function $r$, shared by all agents.
The joint action-value function of $\pi$ is defined as  $Q^\pi(s_t, \mathbf{u}_t) = \mathbb{E}_{s_{t+1:\infty},  \mathbf{u}_{t+1:\infty}}[G_t|s_t, \mathbf{u}_t]$.

%%%%%%%%%%%%%%%%%%%%%%%%%%%%%%%%%%%%%%%%%%%
\section{Method}
%%%%%%%%%%%%%%%%%%%%%%%%%%%%%%%%%%%%%%%%%%%

\begin{figure}[h]
    \centerline{\includegraphics[width=\linewidth]{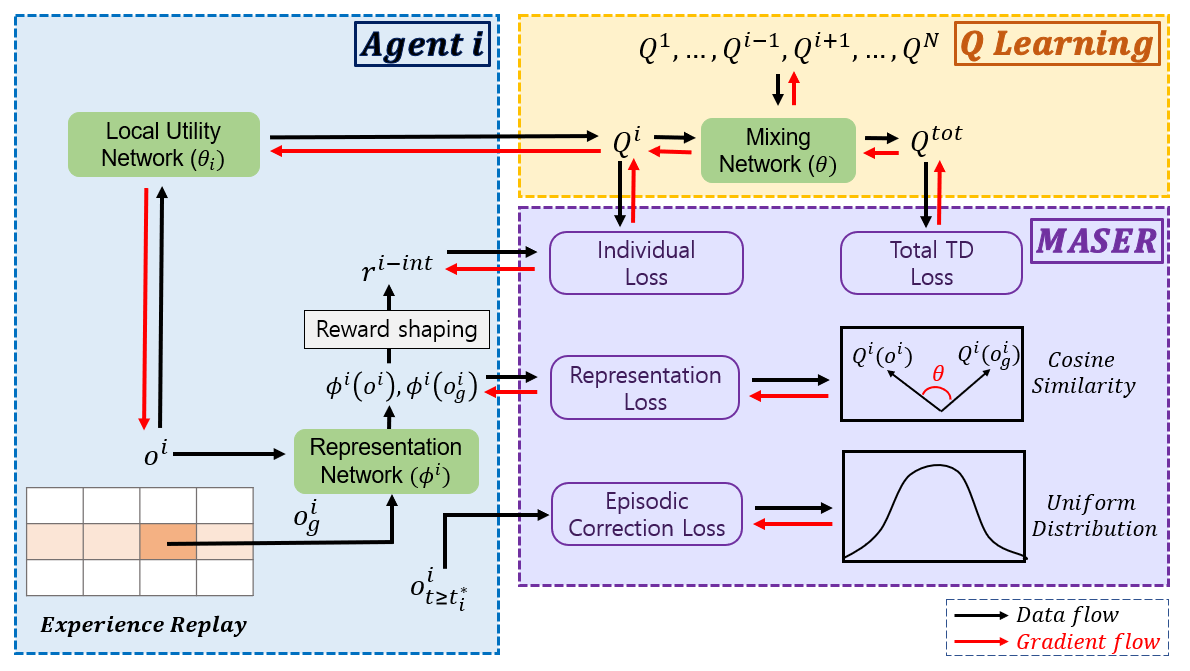}}
    \vskip -0.05in
    \caption{Overflow of MASER:  (Blue box) The blue box shows how MASER assigns subgoals to the agents. MASER  finds out a subgoal for each agent from the replay buffer and computes individual intrinsic rewards from subgoals by using actionable representation learning. (Purple box) The purple box shows the learning loss components of MASER.  From the subgoals from the blue box, it computes total TD loss, individual TD loss, episodic correction loss, and representation learning loss.  (Yellow box) The yellow box shows the Q-learning based on the mixing network.  } 
    \label{fig:overflow}
\end{figure}
\begin{figure*}[h]
    \centerline{\includegraphics[width=\linewidth]{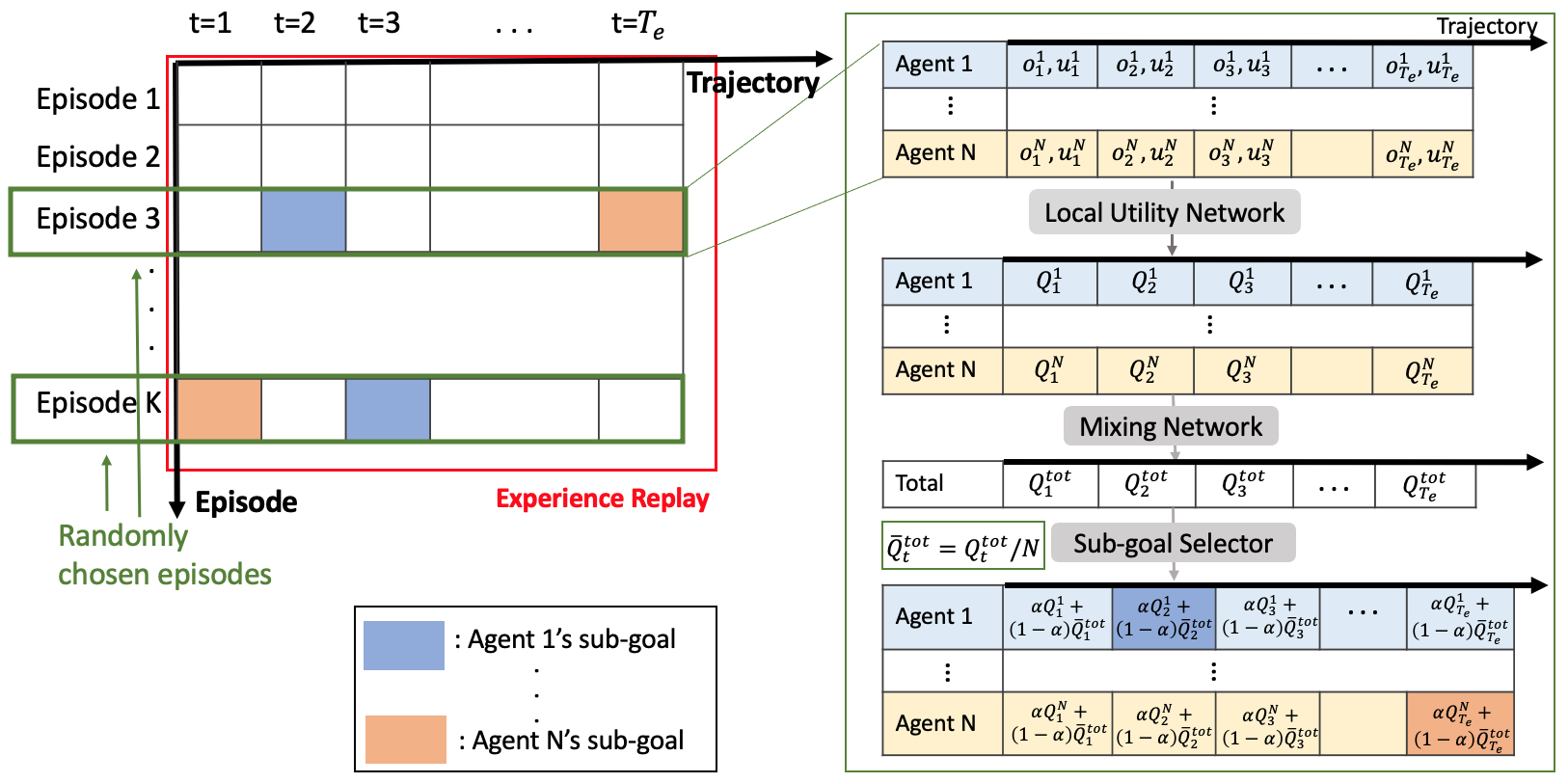}}
    \vskip -0.05in
    \caption{Diagram of subgoal determination from  replay buffer. 
    (Left) MASER chooses $M$ episodes randomly from the  replay buffer and each episode of length $T_e$ has sequential sample information of action and observation of all agents and extrinsic reward  from $t=1$ to $t=T_e$.  (Right) The right side shows how to select a subgoal for each agent based on the current Q-function estimate. }
    \label{fig:subgoal}
    \vspace{2em}
\end{figure*}

In this section, we present MASER solving MARL with sparse reward  effectively based on   subgoals.  We assume the Centralized Training and Decentralized Execution (CTDE) paradigm, which is widely assumed in MARL \cite{2019Oroojlooyjadid, 2018rashid}. We also assume a finite action space $U$ and  episodic off-policy learning in which each episode is composed of $T_e$ time steps.   
The overall architecture of MASER is shown in Fig. \ref{fig:overflow}.  MASER is based on Q-learning and  Q-decomposition similar to QMIX \cite{QMIX}.  That is, 
the overall Q-function network for global return is composed of $N$ local utility networks computing $N$ local Q-values $Q^1,\cdots,Q^N$ (one for each agent) and a mixing network computing the global Q-value denoted as $Q^{tot}$.  At time step $t$, Agent $i$'s local utility network receives the pair of its local observation $o^i_t$ and action $u^i_{t-1}$ to learn the local Q-value $Q^i$. Here, the local utility network incorporates a RNN-type structure to deal with partial observability in $o^i_t$.  Then,  the $N$ local Q-values $Q^1,\cdots,Q^N$  are fed into the mixing network to generate the total Q-value $Q^{tot}$ by enforcing the consistency constraint  
$\frac{\partial Q^{tot}}{\partial Q^i} \ge 0, \forall i$.  In the execution phase after centralized training, the mixing network is removed and Agent $i$ acts based on the local policy derived from the local  action-value function $Q^i$ in a decentralized manner. We implement the mixing network and the local utility networks  by using deep neural networks, and $\theta$ and $\theta_i$ ($i=1,\cdots,N$) denote the network parameters of the mixing network and the $i$-th utility network, respectively.

On the basis of the aforementioned Q-decomposition,  MASER performs   blockwise operation, where the block length is equal to the episode length $T_e$, and adopts the following three key ingredients  to efficiently learn  in environments with sparse reward:

%\vspace{-0.5em}

 \textbf{i) Generating and assigning subgoals}: MASER finds subgoals for agents from the experience replay buffer.  This eliminates the necessity of predesigning good subgoals based on domain knowledge.

%\vspace{-0.5em}

 \textbf{ii) Giving individual rewards}: MASER designs individual rewards for local agents to reach their subgoals while maximizing the joint return.

%\vspace{-0.5em}

\textbf{iii) Actionable distance relevant to Q-learning}: To determine the intrinsic reward based on the Euclidean distance in the transformed domain, MASER uses representational transform based on an actionable distance relevant to Q-learning derived from Amari 0-divergence \cite{Amaribook}.

\begin{comment}

\textbf{iv) Exploration after subgoals}: MASER balances exploitation and exploration. In each episode,  after trying to achieve the subgoals up to a certain time point, MASER performs exploration to search for better states for the final goal.   
\end{comment}

%%%%%%%%%%%%%%%%%%%%%%%%%%%%%%%%%%%%%%%
\subsection{Generating  Subgoals}

As aforementioned, MASER performs blockwise operation, where the block length is equal to the episode length $T_e$. 

Suppose that the current block index is $b$, and  the current mixing network and local utility network  parameters are respectively given by $\theta[b]$ and $\theta_i[b]~(i=1,\cdots,N)$ with  the corresponding Q-functions $Q^{tot}_{\theta[b]}$ and $Q^i_{\theta_i[b]}~(i=1,\cdots,N)$ in the beginning of block $b$.    Suppose also that the  replay buffer stores the episodes up to now, as seen in Fig. \ref{fig:subgoal}.

In order to generate subgoals for block $b$, MASER randomly selects $M$ episodes from the replay buffer.  Each selected episode consists of $\{(o^i_1,u_1^i, \cdots, o^i_{T_e},u^i_{T_e})_{i=1}^N,(r^{ex}_1,\cdots,r^{ex}_{T_e})\}$, where $r^{ex}_t$ is the extrinsic global reward at time step $t$. 
Basically, our choice of the subgoal for each agent is the observation (i.e., partial state) that yields maximum value (i.e., future sum reward) from the selected episode. In order to compute the value of each observation in each of the selected episodes, we exploit the current learned local and global Q-functions  $Q^i_{\theta_i[b]}$ and $Q^{tot}_{\theta[b]}$. 

For explanation, let us consider the $m$-th selected episode. The subgoal for Agent $i$ for block $b$ from the $m$-th selected episode is determined as
\begin{align} 
{t^i_\star} &= \mathop{\arg \max}_{t:1\le t \le T_e} \bigg[\alpha Q^i_{\theta_i[b]} \big(o_t^i, \argmax_{{u^i}}Q^i_{\theta_i[b]}(o_t^i, {u^i})\big)  \nonumber \\ 
& ~~~~~~~~~~~~~~~~~~~ + (1-\alpha) \bar{Q}^{tot}_{\theta[b]}(\mathbf{o}_t, \mathbf{u}_t) \bigg] \nonumber \\
o_g^i  &= o^i_{t^i_\star},  \label{eq:4-subgoal}
\end{align}
where $\alpha \in [0,1]$, $\bar{Q}^{tot}_{\theta[b]}(\mathbf{o}_t, \mathbf{u}_t) = \frac{1}{N} Q^{tot}_{\theta[b]}(\mathbf{o}_t, \mathbf{u}_t)$,  $\mathbf{o}_t = (o_t^1,\cdots,o_t^N)$, $\mathbf{u}_t = (u_t^1,\cdots,u_t^N)$, and  $\mathbf{o}_t$ and  $\mathbf{u}_t$ are from the $m$-th selected episode (for notational simplicity, we omit the episode index $m$).  Note that the subgoal for Agent $i$ for block $b$ is not  the observation that greedily maximizes the local Q-value $Q^i$ based on the current Q-function estimate but is the observation that maximizes the sum of the local Q-value $Q^i$ and the global Q-value $Q^{tot}$ based on the current Q-function estimate.  
The weighting between the two is determined by $\alpha$.  When $\alpha =0$, we have $t^1_\star=\cdots=t^N_\star$ and only the total Q-value is considered with neglection of  individual Q-value variation. 
When $\alpha=1$, on the other hand,  the subgoal for each agent is the greedy local partial state of each agent, and  {$t^i_\star \ne t^j_\star$}  for $i\ne j$ in general in this case. When $0< \alpha < 1$, both  local and global values are considered to determine each agent's subgoal. An ablation study on this aspect is provided in Section 5.
Fig. \ref{fig:subgoal} illustrates the subgoal generation process, where $Q^i_{\theta_i[b]}(o_t^i,u_t^i)$ and $Q^{tot}_{\theta[b]}(\mathbf{o}_t,\mathbf{u}_t)$ are  denoted simply as $Q_t^i$ and $Q^{tot}_t$, respectively. In Fig.  \ref{fig:subgoal},  $o^1_2$ is selected as agent 1's subgoal  and  is $o^N_{T_e}$ is selected as agent $N$'s subgoal for instance.

%%%%%%%%%%%%%%%%%%%%%%%%%%%%%%%%%%%%%%
\subsection{Overall Reward Design}

With the determined  individual subgoals  for all agents for block $b$ from Section 4.1, we now design individual and global rewards to reach subgoals  while maximizing the joint action value. 
We define the individual intrinsic reward for Agent $i$ at time step $t$ as the negative Euclidean distance between the current partial state $o_t^i$ at time step $t$ and the subgoal partial state $o_g^i$ for current block $b$ after representational transform, i.e., \cite{ARC}
\begin{equation} \label{eq:11-intrinsic_loss}
r^{i-int}_t = -||\phi^i(o_t^i) - \phi^i(o_g^i)||_2,~~~t=1,\cdots, T_e,
\end{equation}
\begin{figure}[t]
    \centerline{\includegraphics[width=\linewidth]{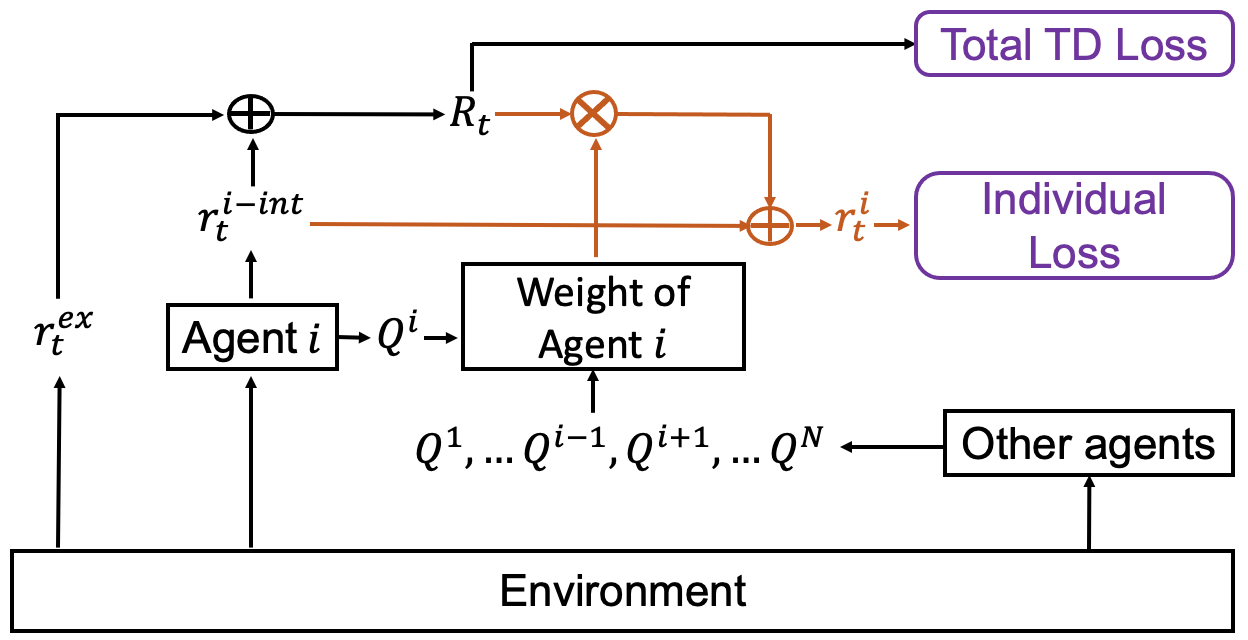}}
    \caption{Overall reward design diagram} 
    \label{fig:reward}
\end{figure}where $\phi^i(\cdot)$ is an actionable representational transform, which will be described in Section 4.3 in detail. By defining intrinsic reward as eq. (\ref{eq:11-intrinsic_loss}) and adding it to the extrinsic reward, each agent tries to reach its  subgoal while maximizing the extrinsic return. Here, we reshape the reward to capture individual contributions to the overall reward. For this, we first define  a proxy reward as the sum of the extrinsic reward from the environment and the intrinsic rewards:
\begin{equation} \label{eq:12-final_reward}
R_t = r_t^{ex} + \lambda \frac{1}{N}\sum_{i=1}^N r^{i-int}_t~~~\mbox{for some~}\lambda >0.
\end{equation}
This proxy reward $R_t$ is used to update  the  mixing network parameter $\theta$. In the proxy reward $R_t$ for updating the mixing network parameter $\theta$, the intrinsic rewards are added so as to make the mixing parameter update nontrivial   at each time step since the extrinsic reward is sparse.  Then, the  individual reward  $r_t^i$ for Agent $i$ used  for updating Agent $i$'s utility network parameter $\theta_i$ is  defined as 
\begin{equation} \label{eq:13-individual_reward}
r_t^i = \underbrace{\frac{\exp(\max_{{u^i}} Q^i_{\theta_i[b]}(o_t^i, {u^i}))}{\sum_{j=1}^N \exp(\max_{{u^j}} Q^j_{\theta_j[b]}(o_t^j, {u^j}))}}_{=:~\text{softmax}\left(\text{max}_{{u^i}}Q^i_{\theta_i[b]}({o_t^i}, {u^i})\right)}\cdot R_t + \lambda r_t^{i-int},
\end{equation}
where  $\text{softmax}(\text{max}_{{u^i}}Q^i_{\theta_i[b]}(o^i_t, {u^i}))$ estimates the contribution of Agent $i$ to the overall reward based on the current Q-function estimate. Note that the first term in the right-hand side (RHS) of eq. (\ref{eq:13-individual_reward}) represents the estimated contribution of Agent $i$ to the overall extrinsic reward $r_t^{ex}$ if there was no intrinsic reward term in $R_t$ in eq. (\ref{eq:12-final_reward}). The intrinsic term $r_t^{i-int}$ is explicitly added again in $r_t^i$ to incorporate achieving the individual subgoal. In this way, both mixing and local utility network parameters are learned to achieve the subgoals as well as maximize the overall extrinsic reward.   The overall reward design  is described in Fig. \ref{fig:reward}.  This reward design process is done for each of the $M$ selected episodes for current block $b$.

%%%%%%%%%%%%%%%%%%%%%%%%%%%%%%%%%%%%%
\subsection{Q-function-based Representation Learning}

% \begin{figure}[h]
%     \centerline{\includegraphics[width=0.8\linewidth]{figure/distance.png}}
%     \caption{Example of why generalization of the subgoal is important. As you can see in this figure, there is wall between the agent (black person) and the goal(red star). If we set the goal as the simple state, the agent can not reach to goal by minimizing the distance between the current state and the goal's state. Therefore, we need to define a novel distance function to provide a general form of goal as the agent can follow with blue arrow. We defined this novel distance function as Q-distance.} 
%     \label{obstacle}
% \end{figure}

As seen in eq. (\ref{eq:11-intrinsic_loss}), we define the individual intrinsic reward $r_t^{i-int}$ for Agent $i$ at time step $t$ as the negative Euclidean distance between the current partial state $o_t^i$ at time step $t$ and the subgoal partial state $o_g^i$  after actionable representation transform $\phi^i$.  An actionable representation can be learned through an actionable distance, which is defined as the difference in actions to reach two different goal states \cite{ARC}. For our Q-learning and Q-decomposition based MARL, we exploit the Q-functions to design an actionable distance between two subgoal states. 
That is, we  consider a soft local policy $\pi^i(\cdot|o_t^i)$ by raising the vector $Q^i_{\theta_i[b]}(o_t^i, \cdot)$ to exponent $e^{\beta(\cdot)}$ and normalizing it, and use the Amari 0-divergence between two distributions $p_1$ and $p_2$, given by \cite{Amaribook}
\begin{equation}  \label{eq:Amari0div}
D^{(0)}(p_1,p_2) = 4\left(1- \sum_u  \sqrt{p_1(u)}\sqrt{p_2(u)} \right),
\end{equation}
which is basically the square of Hellinger distance between the two distributions. The Amari 0-divergence is attractive since it induces a symmetric self-dual Euclidean geometry on the space of distributions (a divergence induces a Riemannian metric,  dual connections, and corresponding geometry \cite{Amaribook}). Furthermore, the nontrivial second term of the RHS of eq. (\ref{eq:Amari0div}) has the nice form of  inner product.  But, this approach requires the determination of an additional temperature parameter $\beta$. So, instead of raising $Q^i_{\theta_i[b]}(o_t^i, \cdot)$ to exponent, normalizing it and taking square root, we directly normalize $Q^i_{\theta_i[b]}(o_t^i, \cdot)$ and put this into inner product. Thus, we define our actionable distance between two subgoal states $o_t^i$ and $o_g^i$ for Q-learning simply as
\begin{equation} \label{eq:5-Q-similar}
\begin{split}
D_{Q}(o_t^i, o_g^i)  
& = 1-~\frac{<Q^i_{\theta_i[b]}(o_t^i, \cdot),  Q^i_{\theta_i[b]}(o_g^i, \cdot)>}{||Q^i_{\theta_i[b]}(o_t^i, \cdot)||\times||Q^i_{\theta_i[b]}(o_g^i, \cdot)||},
\end{split}
\end{equation}
where $Q^i_{\theta_i[b]}(o_t^i, \cdot)$ denotes the vector with the elements of the Q-values for all possible actions for $o_t^i$ with dimension $|U|$, and $<\cdot,\cdot>$ represents the inner product between two vectors. For distance $D_Q$, we only need to compute  the cosine similarity between two vectors. 
Then, we define the loss for learning the representation $\phi^i$ based on the actionable distance $D_Q$ as 
\begin{equation} \label{eq:7-distance_loss}
L_D(\phi^i) = \mathbb{E}_{o_t^i} \Big[ ||\phi^i(o_t^i) - \phi^i(o_g^i)||_2 - D_{Q}(o_t^i, o_g^i) \Big]^2,
\end{equation}
where  $||\cdot||_2$ denotes $L_2$-norm. 
By minimizing the loss $L_D(\phi^i)$, the representational transform $\phi^i$ is learned so that the post-transform Euclidean distance between two partial states becomes similar to one minus  cosine-similarity between the two partial states' Q-value vectors.

\subsection{Overall Learning}

With the designed proxy reward $R_t$ and individual rewards $r_t^i$ for block $b$, local Q and total Q-learning is performed sequentially from time step $t=1$ to $t=T_e$ for block $b$ based on the selected episodes for block $b$. 
The loss function for the $i$-th  local utility network $\theta_i$ is given by 
\[
L_i(\theta_i) = \big[r_t^i + \gamma \text{max}_{{u^i}} Q^i_{\theta_i^-}(o_{t+1}^i, {u^i})  - Q^i_{\theta_i}(o_t^i, u_t^i)\big]^2, \]
where $\theta_i^-$ is the parameter of the $i$-th local target network \cite{dqn}, and $(o_t^i,u_t^i)$ is the $t$-th sample of Agent $i$ in the $m$-th episode selected from the replay buffer at block $b$ (again the episode index $m$ is omitted for notational simplicity).  The loss function for the mixing network parameter $\theta$ is given by
\[
L_{TD}(\theta) = \big[R_t + \gamma \text{max}_{\mathbf{u'}} Q^{tot}_{\theta^-}(s_{t+1}, \mathbf{u'}) - Q^{tot}_{\theta}(s_t,\mathbf{u}_t)\big]^2,
\]
where $\theta^-$ is the parameter of the target network for the mixing network.

\textbf{Episodic Correction:}  Consider the $i$-th agent's sample history  in the $m$-th selected episode: $(o_1^i,u_1^i,o_2^i,u_2^i,\cdots, o_{T_e}^i,u_{T_e}^i)$. Based on the current Q-function estimates $Q_{\theta_i[b]}^i$ and $Q_{\theta[b]}^{tot}$, as a subgoal, we picked the partial state within the sequence $(o_1^i,u_1^i,o_2^i,u_2^i,\cdots, o_{T_e}^i,u_{T_e}^i)$ that maximizes the weighted sum of local Q-value and total Q-value, where the maximizing time step is {$t_\star^i$}, as seen in eq. (\ref{eq:4-subgoal}), and the  proxy reward $R_t$ and individual rewards $r_t^i$ for block $b$ were designed based on this.  Thus, based on the current Q-function estimates $Q_{\theta_i[b]}^i$ and $Q_{\theta[b]}^{tot}$, the subsequence 
$(o_1^i,u_1^i,o_2^i,u_2^i,\cdots, o_{{t^i_\star-1}}^i,u_{{t^i_\star-1}}^i)$ is a proper sequence which increases the Q-value towards the maximum at {$t^i_\star$}. On the other hand, the subsequence 
$(o_{{t^i_\star}}^i,u_{{t^i_\star}}^i, o_{{t^i_\star}+1}^i,u_{{t^i_\star}+1}^i, \cdots, o_{T_e}^i,u_{T_e}^i)$ is an improper sequence according to  the current Q-function estimates $Q_{\theta_i[b]}^i$ and $Q_{\theta[b]}^{tot}$ because this subsequence deteriorates the Q-value away from the maximum.  So, it is highly likely that for $o_t^i$ in this latter subsequence, the (known) action $u_t^i$ is a bad action making transition to a partial state $o_{t+1}^i$ even away from the maximum. Therefore, we want to try diverse actions other than the curren action at $o_t^i$ in this latter subsequence. For this exploration purpose, we apply maximum entropy principle to the timesteps after {$t^i_\star$}. That is, we additionally incorporate the loss function $L_{E,t}(\theta_i)$ for the local utility network parameter $\theta_i$, given by
\begin{equation} \label{eq:Leitthetai}
    L_{E,t}(\theta_i) = D_{KL} \left( \hat{\pi}_t^i(\theta_i)(\cdot|o_t^i) || \pi_U(\cdot|o_t^i)  \right), t={t^i_\star}, \cdots, T_{e},
\end{equation}
where $D_{KL}(\cdot||\cdot)$ is the KL divergence (KLD), $\pi_U$ is the uniform distribution over the action space $U$, and $\hat{\pi}_t^i(\theta_i)$ is a softened $Q$-distribution defined as 
\begin{equation}
\hat{\pi}_t^i(\theta_i)(u|o_t^i) =  \frac{\exp(Q^i_{\theta_i}(o_t^i,{u^i}))}{\sum_{\upsilon^i} \exp(Q^i_{\theta_i}(o_t^i,{\upsilon^i}))},~~~\forall u \in U. \label{eq:pihatitExp}
\end{equation}
Note that minimizing KLD between $ \hat{\pi}_t^i(\theta_i)(\cdot|o_t^i)$ and the uniform distribution is equivalent to maximizing the entropy of $ \hat{\pi}_t^i(\theta_i)(\cdot|o_t^i)$. Note that this extra-loss is due to our way of choosing subgoal based on eq. (\ref{eq:4-subgoal}). Here, in eq. \eqref{eq:pihatitExp} we do not consider a temperature parameter in exponentiation for simplicity. Instead, a weighting factor for this exploration loss is included in the total loss, as seen below.

% \begin{figure}[h]
%     \centerline{\includegraphics[width=0.8\linewidth]{figure/explore.png}}
%     \caption{Exploration after reaching subgoal. In this figure, the black person is the agent and the red star is the final goal. Agent needs to reach the red star, and the black arrow shows the agent's trajectory. Since we designated the subgoal (blue point) as the most important state in the entire trajectory, the trajectory after the subgoal does not have a significant effect on learning. Therefore, SEMARL allows exploration for the trajectory like green arrow after the subgoal.}
%     \label{explore}
% \end{figure}

%%%%%%%%%%%%%%%%%%%%%%%%%%%%%%%%%%%5
The overall learning loss for each of the $m$-th selected episode is given by the sum of total TD loss, individual TD losses, episodic correction losses, and representation losses:
\begin{eqnarray}
&& L(\theta,\theta_i,\phi^i) = L_{TD}(\theta) \label{eq:16-final_loss} \\ 
&&~+ \sum_{i=1}^N \Big[\lambda_I L_i(\theta_i) +\lambda_E\sum_{t={t^i_\star}}^{T_e} L_{E,t}(\theta_i) + \lambda_D L_D(\phi^i)\Big],  \nonumber
\end{eqnarray}
 where $\lambda_I$, $\lambda_E$ and $\lambda_D$ are the weights for the losses.
Finally, the overall learning losses of all $M$ episodes are added and gradient descent to update the parameters  is performed based on the summed learning loss. Once the parameter update is done for block $b$, a new episode is sampled from an $\epsilon$-greedy policy induced from the updated Q-network parameters and is stored into the replay buffer. Then, the learning process for block $b+1$ starts.

%\subsection{Discussion}
%\tcr{Add some intuition or proof why it works well.}

\section{Experiments}

\begin{figure*}[h]
     \centering
     \begin{subfigure}[b]{0.24\textwidth}
         \centering
         \includegraphics[width=\textwidth]{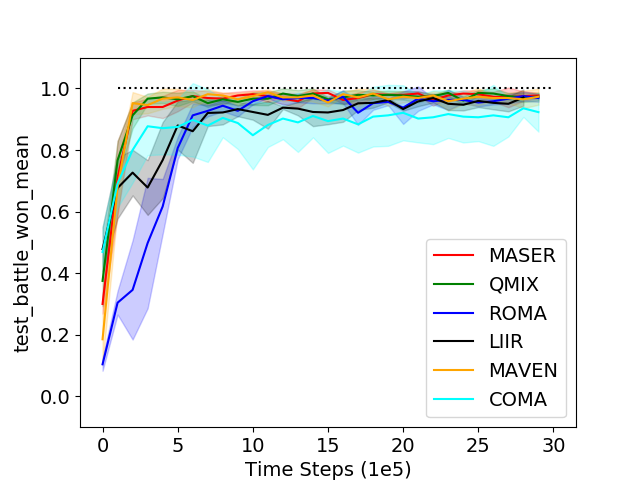}
         \caption{3m (dense)}
         \label{fig:3m_dense}
     \end{subfigure}
     \hfill
     \begin{subfigure}[b]{0.24\textwidth}
         \centering
         \includegraphics[width=\textwidth]{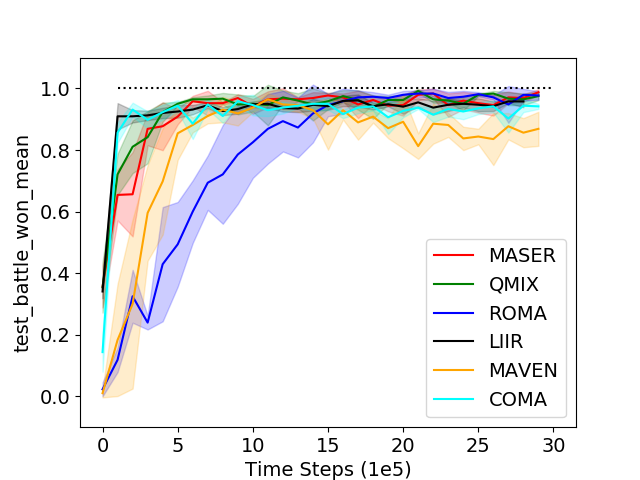}
         \caption{8m (dense)}
         \label{fig:8m_dense}
     \end{subfigure}
     \hfill
     \begin{subfigure}[b]{0.24\textwidth}
         \centering
         \includegraphics[width=\textwidth]{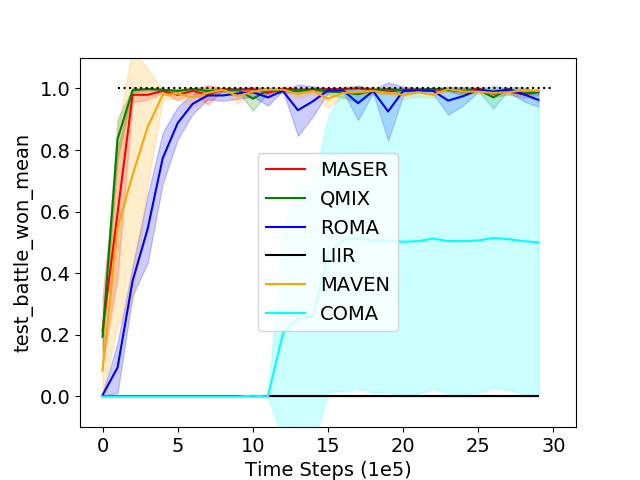}
         \caption{2m\_vs\_1z (dense)}
         \label{fig:2m_vs_1z_dense}
     \end{subfigure}
     \begin{subfigure}[b]{0.24\textwidth}
         \centering
         \includegraphics[width=\textwidth]{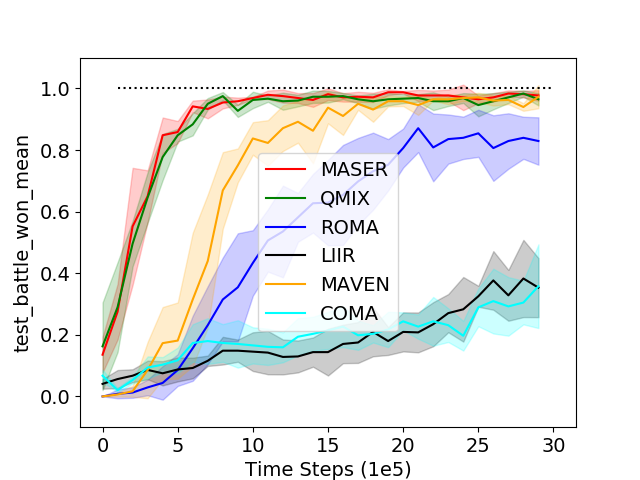}
         \caption{2s3z (dense)}
         \label{fig:2s3z_dense}
     \end{subfigure}
        \caption{Performance of the considered algorithms on the conventional dense-reward setting}
        \label{fig:4map_dense}
\end{figure*}

\begin{figure*}[h]
     \centering
     \begin{subfigure}[b]{0.24\textwidth}
         \centering
         \includegraphics[width=\textwidth]{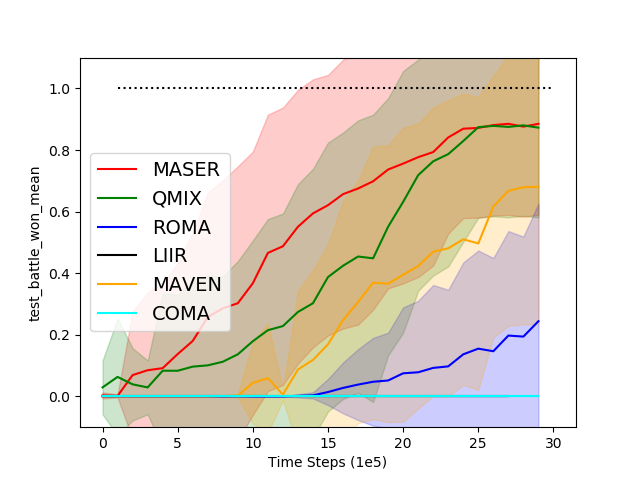}
         \caption{3m (sparse)}
         \label{fig:3m_sparse}
     \end{subfigure}
     \hfill
     \begin{subfigure}[b]{0.24\textwidth}
         \centering
         \includegraphics[width=\textwidth]{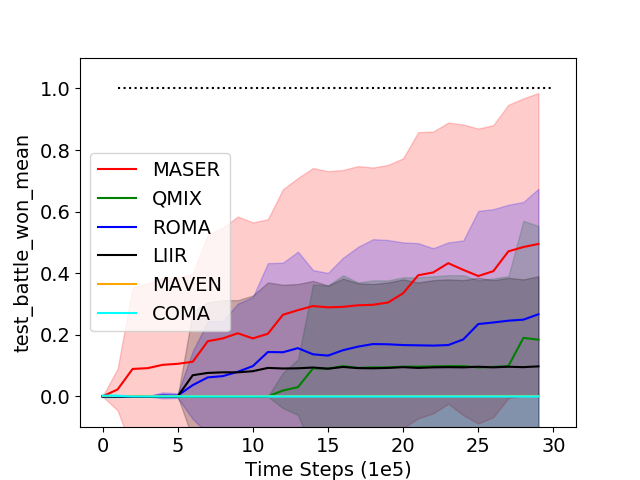}
         \caption{8m (sparse)}
         \label{fig:8m_sparse}
     \end{subfigure}
     \hfill
     \begin{subfigure}[b]{0.24\textwidth}
         \centering
         \includegraphics[width=\textwidth]{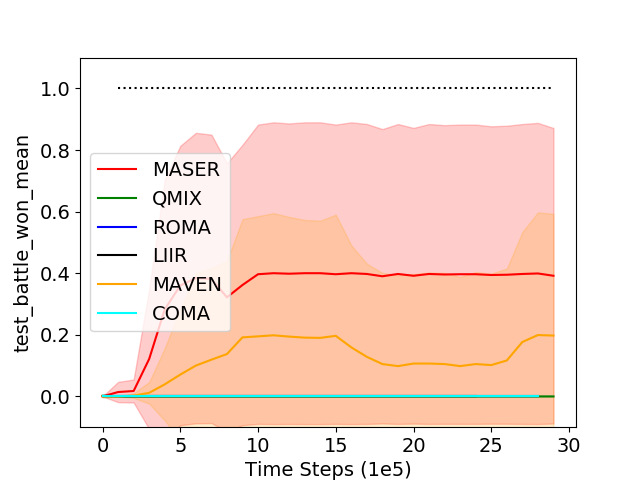}
         \caption{2m\_vs\_1z (sparse)}
         \label{fig:2m_vs_1z_sparse}
     \end{subfigure}
      \begin{subfigure}[b]{0.24\textwidth}
         \centering
         \includegraphics[width=\textwidth]{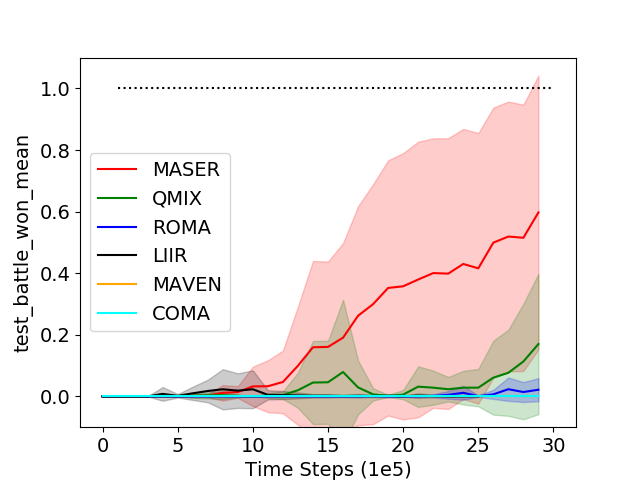}
         \caption{2s3z (sparse)}
         \label{fig:2s3z_sparse}
     \end{subfigure}
        \caption{Performance of the considered algorithms on the sparse-reward  setting}
        \label{fig:4map_sparse}
\end{figure*}

To evaluate  MASER, we considered
the widely-used StarCraft II micromanagement benchmark (SMAC) environment. We conducted experiments on four maps: 3 marines\_vs\_3 marines, 8 marines\_vs\_8 marines, 2 marines\_vs\_1 zealot, and 2 stalkers+3 zealots\_vs\_2 stalkers+3 zealots (denoted as 3m, 8m, 2m\_vs\_1z, 2s3z) in SMAC with dense and sparse reward and compared the performance with  state-of-the-art MARL algorithms: QMIX \cite{QMIX}, ROMA \cite{roma}, LIIR \cite{LIIR}, MAVEN \cite{maven}, and COMA \cite{COMA}. 

\vspace{-1em}
\begin{table}[h]
\centering
\caption{Reward Setting}
\label{tab:reward_sparse_dense}
\resizebox{\columnwidth}{!}
{\begin{tabular}{c|ccc}
\noalign{\smallskip}\noalign{\smallskip}\hline\hline
&  Dense reward & Sparse  reward \\
\hline
All enemies die (Win) & +200 & +200 \\
One enemy dies & +10 & +10  \\
One ally dies  & -5 & -5  \\
Enemy's health  & -Enemy's remaining health  & - \\
Ally's health  & +Ally's remaining health & -  \\
Other elements &  +/- with other components & - \\
\hline
\hline
\end{tabular}}
\end{table}

The reward setting for the dense and sparse reward cases  are shown in Table \ref{tab:reward_sparse_dense}. The dense reward setting is the conventional reward setting in which proper reward is given not only for major events but also for subsidiary side events. To make a sparse-reward StarCraft II setting, we sparsified the rewards of the dense setting as follows:
In the sparse-reward setting, the task is made more difficulty by sparsifying the reward function.  That is, a  reward is given only when one or all enemies die or when one ally dies with no additional reward for state information such as enemy's and ally's health. 
For evaluation, in the case of dense reward, we carried out experiments  with 4 different random seeds  because the dense reward case has lower variance  than the sparse reward case. On the other hand, we carried out
experiments with 10 different random seeds in the sparse case. The result is summarized as the mean of `test battle won mean' which shows the performance in the SMAC environment, and the shaded  area in each figure represents one standard deviation. The hyperparameter setting is available in Appendix A.

\begin{comment}
\tcr{For environment 8m with sparse and dense rewards, $\alpha$ as 0.5, $\lambda$ as 0.03, $\lambda_I, \lambda_D, \lambda_E$ as $0.7*10^{-3}, 0.3*10^{-3}, 0.3*10^{-3}$ respectively.
For environments 3m, 2m\_vs\_1z and 2s3z with sparse and dense rewards, we set $\alpha$ as 0.5, $\lambda$ as 0.03, and $\lambda_I, \lambda_D, \lambda_E$ as $10^{-3}, 10^{-3}, 10^{-3}$ respectively.}
\end{comment}

\begin{figure}[h]
    \begin{subfigure}[b]{0.49\columnwidth}
        \centerline{\includegraphics[width=\linewidth]{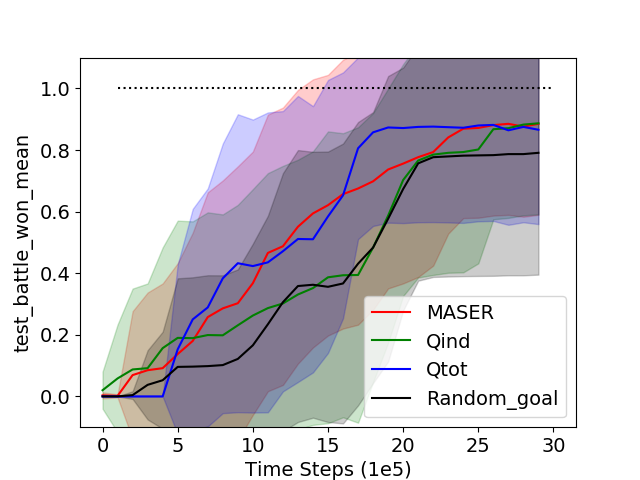}}
        \caption{3m}
     \label{fig:random_goal_3m}
    \end{subfigure}
    \hfill
    \begin{subfigure}[b]{0.49\columnwidth}
        \centerline{\includegraphics[width=\linewidth]{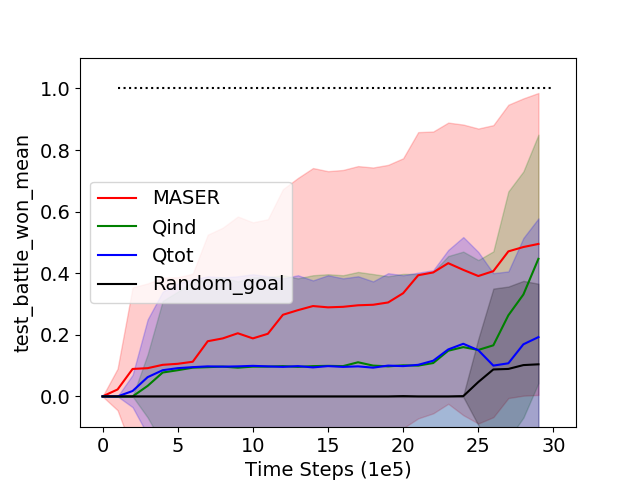}}
        \caption{8m}
     \label{fig:random_goal_8m}
    \end{subfigure}
    \hfill
    \begin{subfigure}[b]{0.49\columnwidth}
        \centerline{\includegraphics[width=\linewidth]{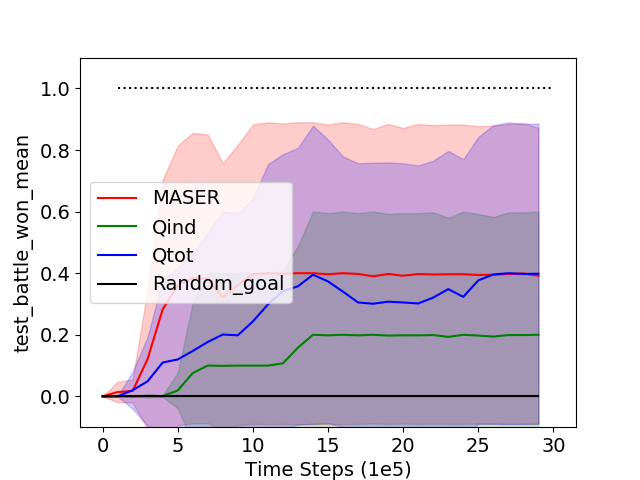}}
        \caption{2m\_vs\_1z}
     \label{fig:random_goal_2m_vs_1z}
    \end{subfigure}
    \hfill
        \begin{subfigure}[b]{0.49\columnwidth}
        \centerline{\includegraphics[width=\linewidth]{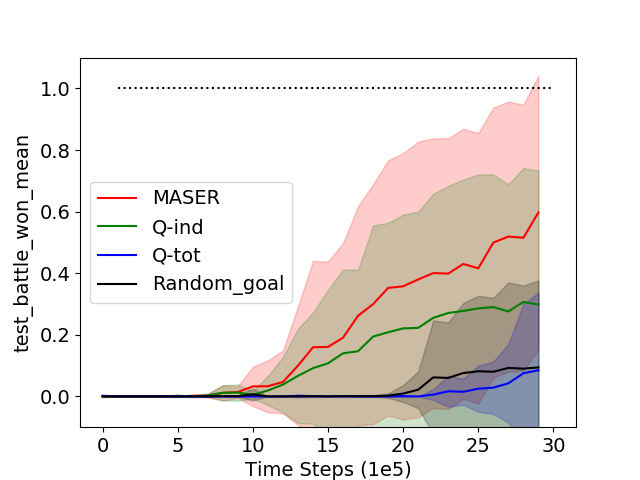}}
        \caption{2s3z}
     \label{fig:random_goal_2s3z}
    \end{subfigure}
    \hfill
    \caption{Performance with respect to different subgoal generation}
    \label{fig:random_goal}
\end{figure}

\begin{figure*}[t]
    \centerline{\includegraphics[width=0.9\textwidth]{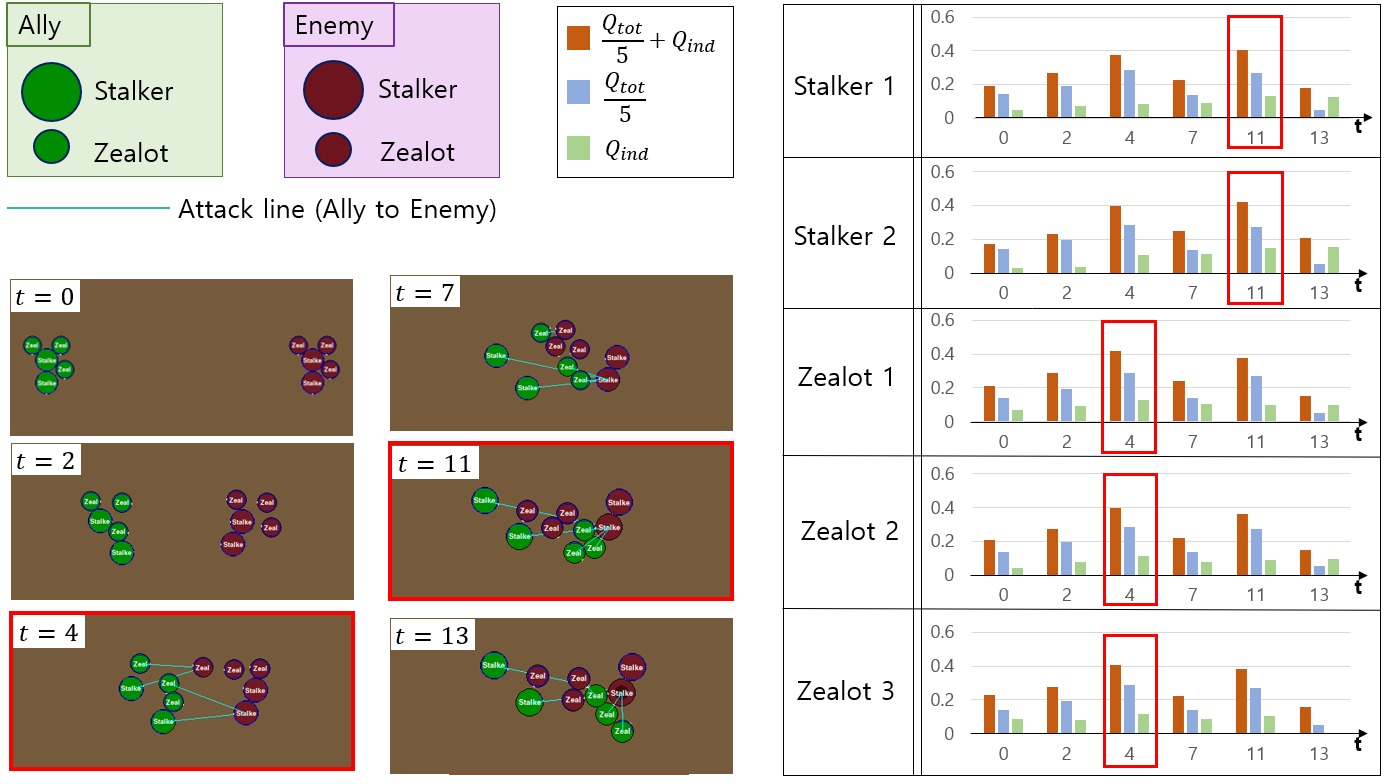}}
    \caption{Selected subgoals for agents: Just before fully learning the task (2s3z). The value that each color indicates in the bar graphs is described in the upper middle region.} 
    \label{fig:subgoalIllust}
\end{figure*}

\textbf{Performance on StarCraft II Micromanagement:}    We first ran all the considered algorithms on the conventional dense-reward setting and the result is shown in Fig. \ref{fig:4map_dense}. It is seen that most of the algorithms work well in 3m, 8m, 2m\_vs\_1z, and some algorithms are worse than others in 2s3z.  We then ran the algorithms on the difficult sparse-reward setting and the result is shown in  Fig. \ref{fig:4map_sparse}.  It is seen that MASER significantly outperforms  other state-of-the-art algorithms: roll-based ROMA, intrinsic reward learning LIIR, exploration-based MAVEN in the difficult sparse-reward multi-agent setting. 
In the relatively easy sparse 3m task, QMIX and MAVEN can also learn the task, as seen in 
Fig. \ref{fig:3m_sparse}. In the more difficult sparse 8m and 2m\_vs\_1z tasks, however, other algorithms except MASER have difficulty in learning the tasks, as seen in Figs.  \ref{fig:8m_sparse} and  \ref{fig:2m_vs_1z_sparse}.  We tested MASER on a heterogeneous environment 2s3z. It is seen that MASER significantly outperforms other algorithms.

\textbf{Subgoal Generation:}   In order to  see the effectiveness of our subgoal design, we compared MASER with the method of choosing random subgoals for agents  from the  replay buffer. We also compared with the case in which the subgoal is selected based only on individual Q-value $Q^i$ or only on total Q-value $Q^{tot}$ (MASER considers both individual and total Q-values by setting $\alpha=0.5$ in eq. (\ref{eq:4-subgoal}).) The comparison result on several StarCraft II micromanagement tasks  are shown in Fig.   \ref{fig:random_goal}. It is seen that considering Q-values performs better than the random selection, and  MASER considering both individual and total Q-values performs best.  
We recognize from Fig. \ref{fig:random_goal}  that  considering individual Q-values is important. To see the reason, we investigated how subgoals for agents for the 2s3z task were selected by MASER. In 2s3z,  domain knowledge tells us that it is advantageous  that stalkers with weak health perform long-distance attacks and zealots with high health perform short-distance attacks. Fig. \ref{fig:subgoalIllust} shows the selected subgoals for all five agents just before fully learning the task (around 250M timesteps). The selected subgoals for stalkers are the partial state at ${t^i_\star}=11$ and the selected subgoals for zealots are the partial states at   ${t^i_\star}=4$ from the chosen episode from the buffer
(for ${t^i_\star}$, please see eq. \eqref{eq:4-subgoal}). The corresponding situations are shown in the left figures.  In the figure at $t=4$, we can see that zealots with high health are ahead of stalkers with weak health, protecting stalkers and preparing for short-distance attacks. In the figure at $t=11$, we can see that stalkers are helping zealots from a long distance while multiple zealots are attacking a single enemy stalker. Both situations are advantageous in 2s3z from our domain knowledge. We can observe that by considering both individual and total Q-values, MASER leads to certain role assignments like in ROMA, while achieving the global goal together.

\begin{figure}[h]
     \centering
     \begin{subfigure}[b]{0.45\columnwidth}
         \centering
         \includegraphics[width=\textwidth]{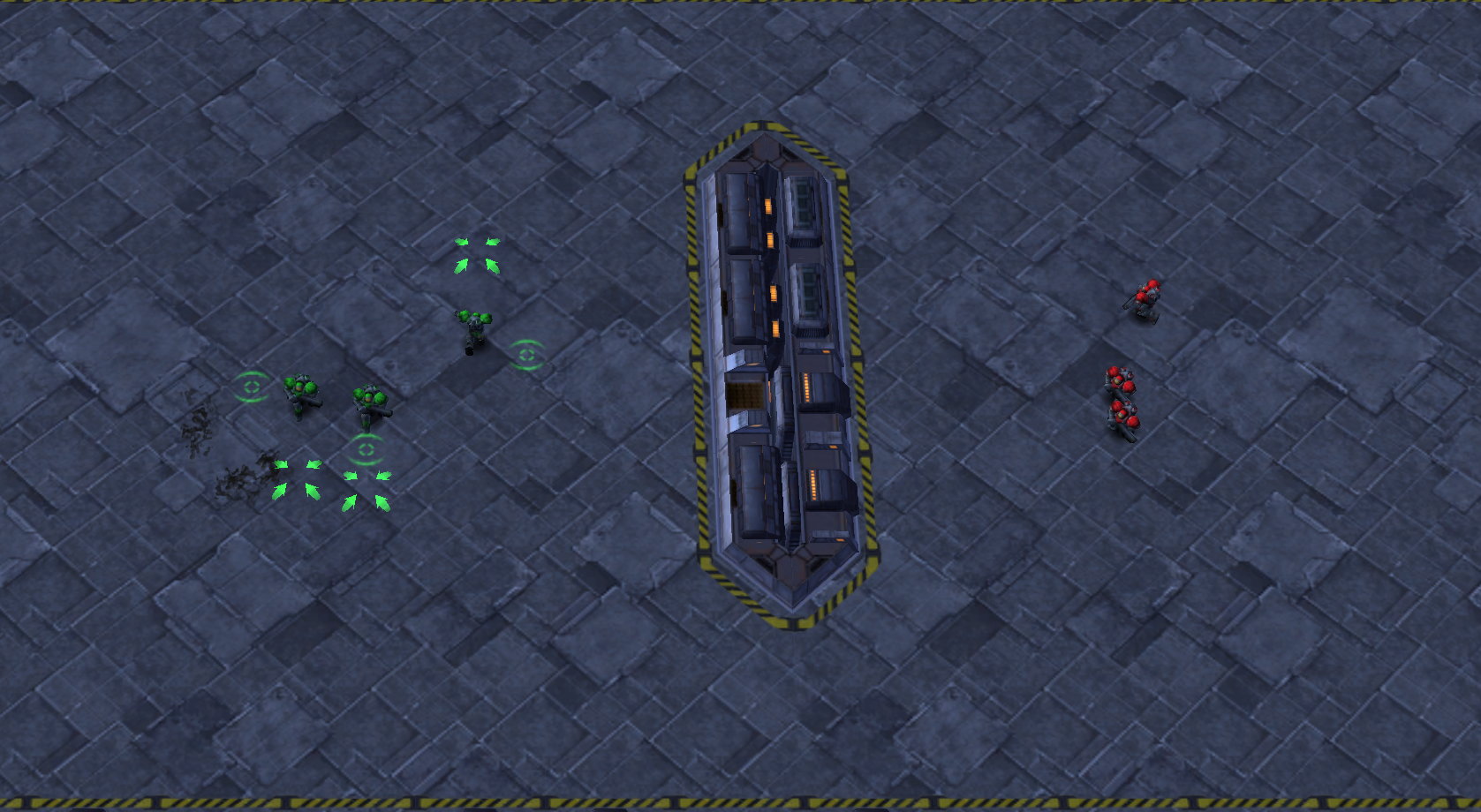}
         \vspace{0.03cm}
         \caption{3m-cliff environment}
         \label{fig:3m_cliff_envrionment}
     \end{subfigure}
     \hfill
     \begin{subfigure}[b]{0.53\columnwidth}
         \centering
         \includegraphics[width=\textwidth]{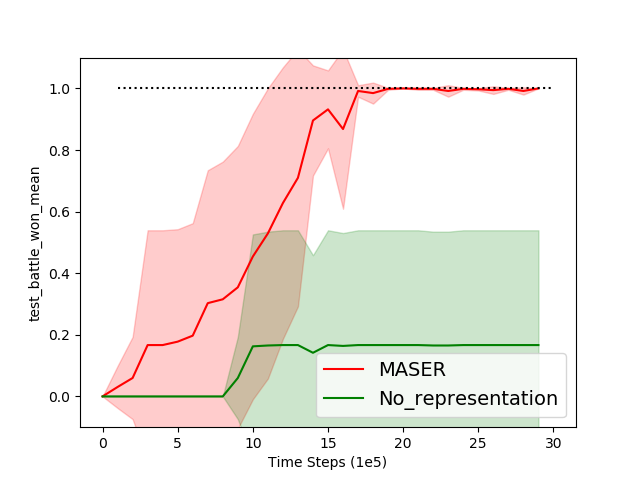}
         \caption{3m-cliff}
         \label{fig:3m_cliff_result}
     \end{subfigure}
           \caption{Impact of  Q-function-based actionable representation}
        \label{fig:3m_cliff}
\end{figure}

\textbf{Q-function-based Actionable Distance:} We tested our representation learning based on the proposed actionable distance using Q-function. For this, we created an environment, given by 3m with a cliff with sparse reward, as shown in Fig. \ref{fig:3m_cliff_envrionment}, which has a blocking cliff.  Fig.          \ref{fig:3m_cliff_result} shows the performance of MASER in this environment, averaged over  6 different random seeds.      Fig. \ref{fig:3m_cliff_result} shows the result. It is seen that MASER with the intrinsic reward in eq. \eqref{eq:11-intrinsic_loss} learning the  representation $\phi^i$ based on Q-function effectively learns the task, while the baseline with no representation learning mostly fails in learning.

\begin{figure}[t]
    \begin{subfigure}[b]{0.49\columnwidth}
        \centerline{\includegraphics[width=\linewidth]{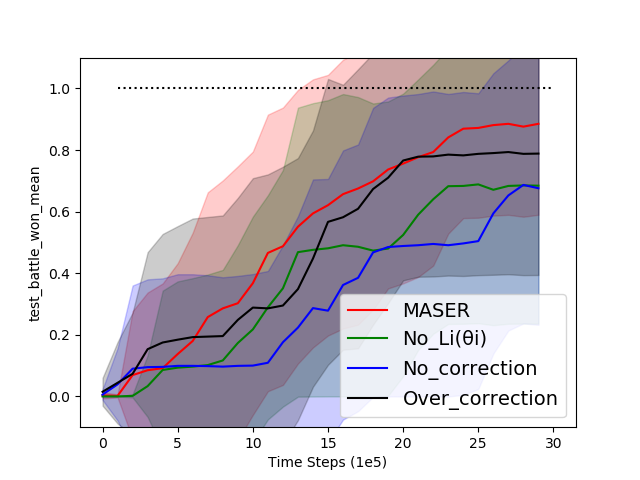}}
        \caption{3m}
     \label{fig:ablation_3m}
    \end{subfigure}
    \hfill
    \begin{subfigure}[b]{0.49\columnwidth}
        \centerline{\includegraphics[width=\linewidth]{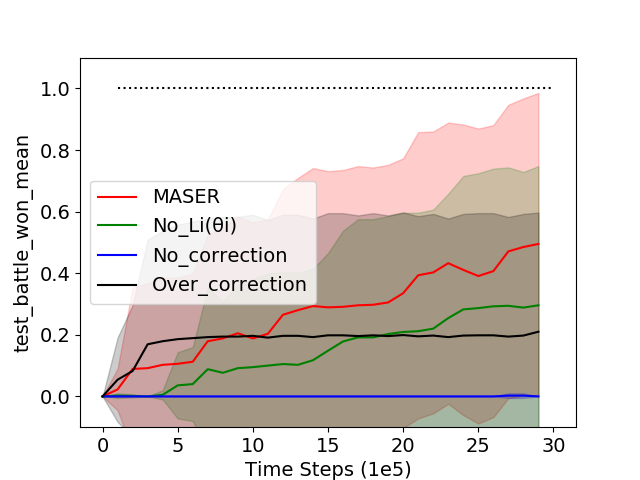}}
        \caption{8m}
     \label{fig:ablation_8m}
    \end{subfigure}
    \hfill
    \begin{subfigure}[b]{0.49\columnwidth}
        \centerline{\includegraphics[width=\linewidth]{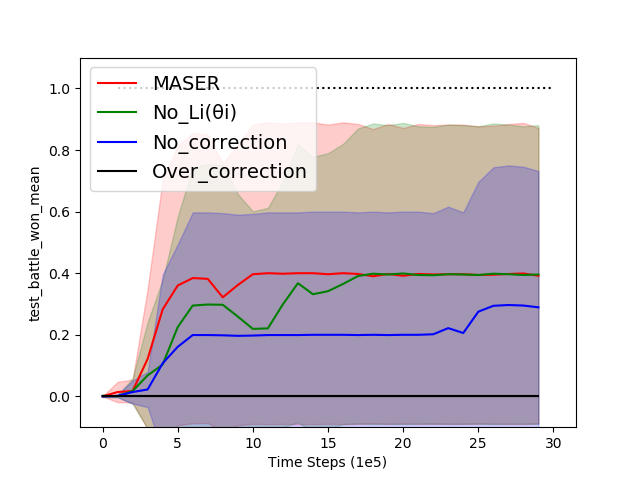}}
        \caption{2m\_vs\_1z}
     \label{fig:ablation_2m_vs_1z}
    \end{subfigure}
    \hfill
        \begin{subfigure}[b]{0.49\columnwidth}
        \centerline{\includegraphics[width=\linewidth]{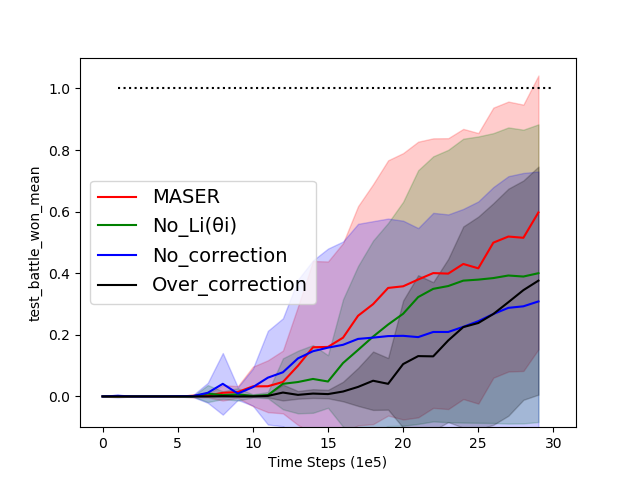}}
        \caption{2s3z}
     \label{fig:ablation_2s3z}
    \end{subfigure}
    \hfill
    \caption{Performance regarding individual loss and episodic correction}
    \label{fig:each_component}
\end{figure}

\textbf{Further Ablation Study:}  For further ablation study, we considered the following :

- No loss function $L_i(\theta_i)$ for individual Q update in eq. \eqref{eq:16-final_loss}: In this case, the parameters of the local utility networks are updated solely by back propagation from $L_{TD}(\theta)$. This case is denoted as `No\_Li($\theta i$) in Fig.\ref{fig:each_component}.

- No episodic correction mentioned in Sec. 4.4. This case is denoted as `No correction' in Fig. \ref{fig:each_component}.

- Episodic correction from the start of the block: In this case, the reward design is the same, but the loss in eq. \eqref{eq:Leitthetai} starts from $t=1$ not ${t^i_\star}$. This case is denoted as `over correction' in Fig.  \ref{fig:each_component}.

The results on StarCraft II micromanagement tasks `3m', `8m', `2m\_vs\_1z' and `2s3z' with these components ablated are shown in  Fig. \ref{fig:each_component}. It is seen that each component of MASER contributes to the overall good performance of MASER.

The source code of the proposed algorithm is available at \url{https://github.com/Jiwonjeon9603/MASER}.

\section{Conclusion}
In this paper, we have introduced a novel method of assigning subgoals from the experience replay buffer for  sparse-reward multi-agent RL environments.  We have chosen the subgoals for multiple agents as the partial states in episodes selected from the replay buffer, maximizing the mixed individual and total Q-values. This choice enables the agents to learn behavior optimizing both individual roles and the global goal. Using the chosen subgoals, we have generated intrinsic reward based on  representation learning using a novel Q-function-based actionable distance  and designed an overall reward architecture for sparse-reward multi-agent RL.  Finally, we have demonstrated the effectiveness of the proposed MARL learning scheme with StarCraft II micromanagement tasks, and experiment results show its good performance in the considered sparse-reward MARL tasks. 

\section*{Acknowledgments}
This work was conducted by Center for Applied Research in Artificial Intelligence (CARAI) grant funded by Defense Acquisition Program Administration (DAPA) and Agency for Defense Development (ADD) (UD190031RD).

% Acknowledgements should only appear in the accepted version.
% \section*{Acknowledgements}

% \textbf{Do not} include acknowledgements in the initial version of
% the paper submitted for blind review.

% If a paper is accepted, the final camera-ready version can (and
% probably should) include acknowledgements. In this case, please
% place such acknowledgements in an unnumbered section at the
% end of the paper. Typically, this will include thanks to reviewers
% who gave useful comments, to colleagues who contributed to the ideas,
% and to funding agencies and corporate sponsors that provided financial
% support.

% In the unusual situation where you want a paper to appear in the
% references without citing it in the main text, use \nocite
\nocite{langley00}

\bibliography{maser}
\bibliographystyle{icml2022}

%%%%%%%%%%%%%%%%%%%%%%%%%%%%%%%%%%%%%%%%%%%%%%%%%%%%%%%%%%%%%%%%%%%%%%%%%%%%%%%
%%%%%%%%%%%%%%%%%%%%%%%%%%%%%%%%%%%%%%%%%%%%%%%%%%%%%%%%%%%%%%%%%%%%%%%%%%%%%%%
% APPENDIX
%%%%%%%%%%%%%%%%%%%%%%%%%%%%%%%%%%%%%%%%%%%%%%%%%%%%%%%%%%%%%%%%%%%%%%%%%%%%%%%
%%%%%%%%%%%%%%%%%%%%%%%%%%%%%%%%%%%%%%%%%%%%%%%%%%%%%%%%%%%%%%%%%%%%%%%%%%%%%%%
\newpage
\appendix
\onecolumn
\section{Details in Implementation}

We built MASER on top of QMIX \cite{QMIX}. Our code is based on Pytorch and we used NVIDIA-TITAN Xp. The values of hyper-parameters are shown in Table \ref{hyperparameter}. 

In MASER, the most recent 5000 episodes are stored in the replay buffer, and the mini-batch size is 32. MASER updates the target network every 200 episodes. We used RMSProp for the optimizer and the learning rate of the optimizer is 0.0005. The discounted factor of expected reward (i.e. return) is 0.99. And the value of epsilon for epsilon-greedy Q-learning starts at 1.0 and ends at 0.05 with 50000 anneal time. During the training, the maximum time step is 3005000 in all experiments. We carried out 10 random seeds for the sparse environments, 4 random seeds for the dense environment, and 6 random seeds for the `3m with cliff' environment.

The agent network is a DRQN \cite{drqn} which  consists  of GRU \cite{gru} for the recurrent layer with a 64-dimensional hidden layer, and 64-dimensional fully-connected layer before and after the GRU layer. For the mixing network, the embedding dimension is 32. For the representation network, we used a neural network with 1 hidden layer with ReLU activation function and 1 output layer with a linear function. The dimension of an input layer, hidden layer, and output layer was the observation dimension, 128, and the action dimension, respectively.

The hyper-parameters for MASER are as follows. For `8m environment', $\alpha$ is 0.5, $\lambda$ is 0.03, and $\lambda_I$, $\lambda_D$ and $\lambda_E$ are 0.0007, 0.0003 and 0.0003, respectively. All experiments except for `8m environment', $\alpha$ is 0.5, $\lambda$ is 0.03, and $\lambda_I$, $\lambda_D$, $\lambda_E$ are all 0.001 respectively.

\begin{table}[h!]
\centering
\caption{Hyper-parameters of MASER and five baselines for SMAC tasks.}
\label{hyperparameter}
\begin{tabular}{c|cccccc}
\noalign{\smallskip}\noalign{\smallskip}\hline\hline
& MASER & QMIX & ROMA & LIIR & MAVEN & COMA \\
\hline
Replay buffer size & 5000 & 5000 & 5000 & 32 & 5000 & 8 \\
Mini-batch size & 32 & 32 & 32 & 32 & 32 & 8 \\
Optimizer & RMSProp & RMSProp & RMSProp & RMSProp & RMSProp & RMSProp \\
Agent Runner & episode & episode & parallel & parallel & episode & parallel \\
Learning rate & 0.0005 & 0.0005 & 0.0005 & 0.0005 & 0.0005 & 0.0005 \\
Discounted factor & 0.99 & 0.99 & 0.99 & 0.99 & 0.99 & 0.99 \\
Epsilon anneal step & 50000 & 50000 & 50 & 50000 & 50000 & 100000 \\
Target update interval & 200 & 200 & 200 & 200 & 200 & 200 \\
Mixing network dimension & 32 & 32 & 32 & - & 32 & - \\

\hline
\hline
\end{tabular}
\end{table}

%%%%%%%%%%%%%%%%%%%%%%%%%%%%%%%%%%%%%%%%%%%%%%%%%%%%%%%%%%%%%%%%%%%%%%%%%%%%%%%
%%%%%%%%%%%%%%%%%%%%%%%%%%%%%%%%%%%%%%%%%%%%%%%%%%%%%%%%%%%%%%%%%%%%%%%%%%%%%%%

\end{document}